%% file: acl_latex.tex
\documentclass[11pt]{article}

\usepackage{amsthm, amsmath, amssymb}
\usepackage{mathrsfs}

\usepackage[preprint]{acl}

\usepackage{times}
\usepackage{latexsym}
\usepackage{algorithm}
\usepackage{algorithmicx}
\usepackage{algpseudocode}
\usepackage{booktabs}
\usepackage{multirow}
\usepackage[table]{xcolor}
\usepackage{makecell}
\usepackage[most]{tcolorbox}
\usepackage{listings}
\definecolor{promptcolor}{RGB}{237,242,236}
\definecolor{promptcolorheader}{RGB}{199,216,196}
\definecolor{datacolor}{RGB}{235,241,249}
\definecolor{datacolorheader}{RGB}{194,211,233}
\lstdefinestyle{prompt}{
  basicstyle=\ttfamily\scriptsize,
  breaklines=true,
  breakatwhitespace=false,
  frame=none,
  backgroundcolor=\color{promptcolor!35},
  xleftmargin=8pt,
  xrightmargin=8pt,
  aboveskip=6pt,
  belowskip=4pt,
  keepspaces=true,
  showstringspaces=false,
}
\newtcolorbox{promptbox}[2][]{
  enhanced,
  breakable,
  colback=promptcolor!35,
  colframe=promptcolorheader,
  boxrule=0.5pt,
  arc=1pt,
  left=0.5em,
  right=0.5em,
  top=0.4em,
  bottom=0.4em,
  title={#2},
  fonttitle=\bfseries\small,
  colbacktitle=promptcolorheader,
  coltitle=black,
  #1
}
\newtcolorbox{databox}[2][]{
  enhanced,
  breakable,
  colback=datacolor!55,
  colframe=datacolorheader,
  boxrule=0.5pt,
  arc=1pt,
  left=0.5em,
  right=0.5em,
  top=0.4em,
  bottom=0.4em,
  title={#2},
  fonttitle=\bfseries\small,
  colbacktitle=datacolorheader,
  coltitle=black,
  #1
}

\usepackage{tablefootnote}
\usepackage{subcaption}
\usepackage{tikz}
\usepackage{pgfplots}
\usepackage{enumitem}
\pgfplotsset{compat=1.18}

\usepackage[T1]{fontenc}

\usepackage[utf8]{inputenc}

\usepackage{microtype}

\usepackage{inconsolata}

\usepackage{graphicx}

\newcommand{\ourmethod}{SVR}

\title{Support Vector Rubrics: \\Closing the Gap Between Self-Generated and Human Rubrics}

\author{
Mengyuan Sun\textsuperscript{1}\thanks{Equal contribution.},
Yu Li\textsuperscript{2}\footnotemark[1],
Zhuohao Yu\textsuperscript{1},
Shikun Zhang\textsuperscript{1},
Wei Ye\textsuperscript{1}\thanks{Corresponding author.} \ \\
\textsuperscript{1}National Engineering Research Center for Software Engineering, Peking University \ \\
\textsuperscript{2}University of Science and Technology of China \ \\
\texttt{{mengyuansun25}@stu.pku.edu.cn},
\texttt{liyu01@mail.ustc.edu.cn},
\texttt{wye@pku.edu.cn}
}

\begin{document}
\maketitle
\begin{abstract}

Rubric-based evaluation is a promising paradigm for judging large language model (LLM) outputs, yet self-generated rubrics lag human-annotated criteria on hard instances.
We argue this \emph{discriminative gap} reflects an objective mismatch: self-generated rubrics \emph{describe} good responses, whereas effective criteria must \emph{discriminate} between close candidates.
To close this gap, we introduce \textbf{\ourmethod{}} (\textsc{Support Vector Rubrics}), a framework that recasts rubric construction as max-margin boundary learning over preference data. \ourmethod{} mines contrastive features from preference pairs into a rubric bank, learns a prompt-conditioned selector together with global rubric weights, and iteratively refines the bank through support-pair selection and adversarial probing of hard negatives.
At inference, given only the prompt, \ourmethod{} retrieves the top-$k$ rubrics from the bank and scores responses.
On RubricBench, \ourmethod{} narrows the gap to human reference rubrics from \textbf{24.1} to \textbf{0.3} points and outperforms strong self-rubric and judge baselines, and the learned bank transfers across judges without retraining. On RewardBench~1\&2, and RM-Bench, it remains competitive with dedicated reward models, demonstrating broader reward modeling capability.
Overall, boundary-defining rubrics offer a principled route to closing the discriminative gap in LLM evaluation.

\end{abstract}

\input{sec1-intro}

\input{sec2-related}
\input{sec3-methodology}
\input{sec4-experiments}
\input{sec5-analysis}

\input{sec5-conclusion}

\section*{Limitations}

We train the rubric bank on preference data from a single LLM judge (GPT-OSS-120B) and do not incorporate human annotations, multi-judge ensembles, or distilled lightweight judges, so the bank inherits whatever systematic biases the training judge carries.
Although our method is motivated by max-margin classifier, the rubric bank is a discrete set of natural-language criteria without a canonical kernel feature map, so $\alpha(x)$ is realised as a prompt-conditioned MLP and the convergence guarantees and hyperparameter selection rules of kernelised SVR do not transfer rigorously to our setting.
While the trained margin can in principle serve as a dense reward signal for reinforcement learning on open-ended generation tasks, our experiments only verify its discrimination quality on preference pairs and do not directly demonstrate that rubric-level gains transfer downstream to policy model optimisation.

\bibliography{custom}

\clearpage

\input{appendix}

\end{document}

%% file: sec1-intro.tex
\section{Introduction}
\label{sec:intro}

\begin{figure}[t]
\centering
\includegraphics[width=\columnwidth]{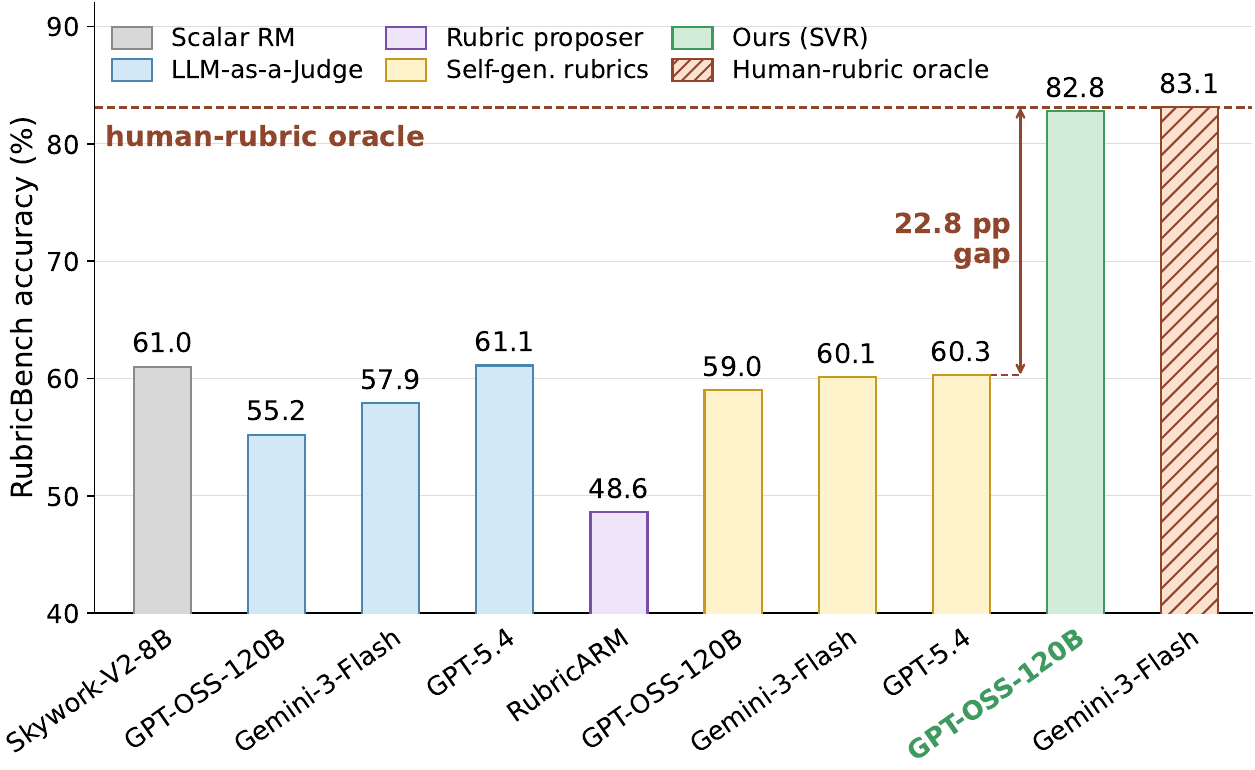}
\caption{\textbf{\textbf{The discriminative gap on RubricBench.}} LLM self-generated rubrics, state-of-the-art scalar reward model, and frontier LLM all stall below $62$, far short of the human-rubric oracle at $83.1$. \ourmethod{} closes this gap, lifting GPT-OSS-120B to $82.8$.}
\label{fig:gap}
\end{figure}

Aligning large language models with human preferences hinges on reliable evaluation~\citep{NIPS2017_d5e2c0ad, NEURIPS2020_1f89885d}, yet open-ended tasks such as question answering and general helpfulness are inherently subjective, and cannot be adequately captured by binary correctness or opaque scalar rewards~\citep{zheng2023judging,li2025generation}.
Generative reward models (GRMs) introduce chain-of-thought judgments~\citep{zhang2025generative, chen2026rmr, yu2025rewardanything}, but their free-form criteria remain prone to superficial presentation bias~\citep{liu2025rm,zhang2026rubricbench} and reward hacking~\citep{coste2024reward,gao2023scaling}.
Rubric-based evaluation closes both gaps by decomposing quality into structured, verifiable criteria that align model judgments with human preferences~\citep{gunjal2026rubrics, liu2026xpertbench}.
An increasingly common practice is to let the judge model write the rubric on the fly from the prompt~\citep{liu2025openrubrics,cook2024ticking,lee2025checkeval}, which scales without human annotation but leaves a basic question open: \textit{are these self-generated rubrics actually reliable?}

Recent evidence suggests \textit{they are not}.
On RubricBench~\citep{zhang2026rubricbench}, a pairwise benchmark deliberately filtered for nuance and surface-level biases, replacing self-generated rubrics with human-annotated ones lifts pairwise accuracy by more than $20$ percentage points with the judge model held fixed, and the gain is stable across backbones.
Figure~\ref{fig:gap} illustrates this \emph{discriminative gap}.
We argue the gap is one of \emph{objective}, not language ability: self-generated rubrics default to \emph{describing} good responses with generic dimensions such as clarity and completeness that hold uniformly across answers, whereas human rubrics \emph{discriminate} responses near the decision boundary with specific factors, such as whether a safety caveat is present or a tempting counter-example is correctly handled.
\citet{shen2026rethinking} corroborate this diagnosis, attributing the failure of prompt-only rubrics to insufficient coverage of nuanced quality dimensions, misaligned or overlapping criteria, and correlated items that double-count the same signal.
This framing of rubrics as boundary-relevant criteria has a natural parallel in classification theory.
The classical lesson from max-margin classifiers~\citep{boser1992training,cortes1995support,bartlett1998boosting} is that a decision boundary is shaped by \emph{support vectors}, and by analogy an effective rubric should be tuned to the support pairs of the response space, the hard cases near the boundary, instead of to its centroid.

Building on this view, we propose \ourmethod{} (\textsc{Support Vector Rubrics}), which reframes rubric construction as max-margin boundary learning over preference data.
Instead of writing a fresh rubric per prompt, \ourmethod{} maintains a \emph{global rubric bank} contrastively mined from preference pairs, and learns a prompt-conditioned selector $\alpha(x)$ together with a global importance weight $w$ per rubric that define a discriminative margin between chosen and rejected responses.
The bank is refined over rounds: samples near or on the wrong side of the margin are designated as \emph{support pairs}, an adversarial probe synthesises \emph{hard negatives} against them, and a contrastive induction step adds rubrics for the newly exposed failure modes before pruning low-weight, low-support, or redundant entries.
At inference, \ourmethod{} retrieves the top-$k$ rubrics from the learned bank conditioned only on the prompt, fixing the evaluation criterion before any candidate response is observed. This decouples rubric generation from response content, providing a principled and reusable standard for downstream scoring, ranking, and reinforcement learning from human feedback (RLHF) pipelines.

\paragraph{Contributions.} (i) We reframe the \emph{discriminative gap} as a mismatch between descriptive and discriminative objectives at the rubric proposal stage, showing that the judge model can score reliably once given adequate criteria. (ii) We propose \ourmethod{}, which constructs rubrics by max-margin boundary learning over a global bank with a prompt-conditioned selector, learned importance weights, and adversarial refinement. (iii) \ourmethod{} reaches $82.8$ on RubricBench, within $0.3$ of the human-rubric oracle, while staying competitive on RewardBench~1\&2 and RM-Bench. (iv) The learned bank is trained once and transfers across both heterogeneous judge families and unseen preference benchmarks, requiring no human rubric annotation and no per-prompt rubric generation at inference.

%% file: sec2-related.tex
\section{Related Work}
\label{sec:related}

\begin{figure*}[t]
\centering
\includegraphics[width=0.98\textwidth]{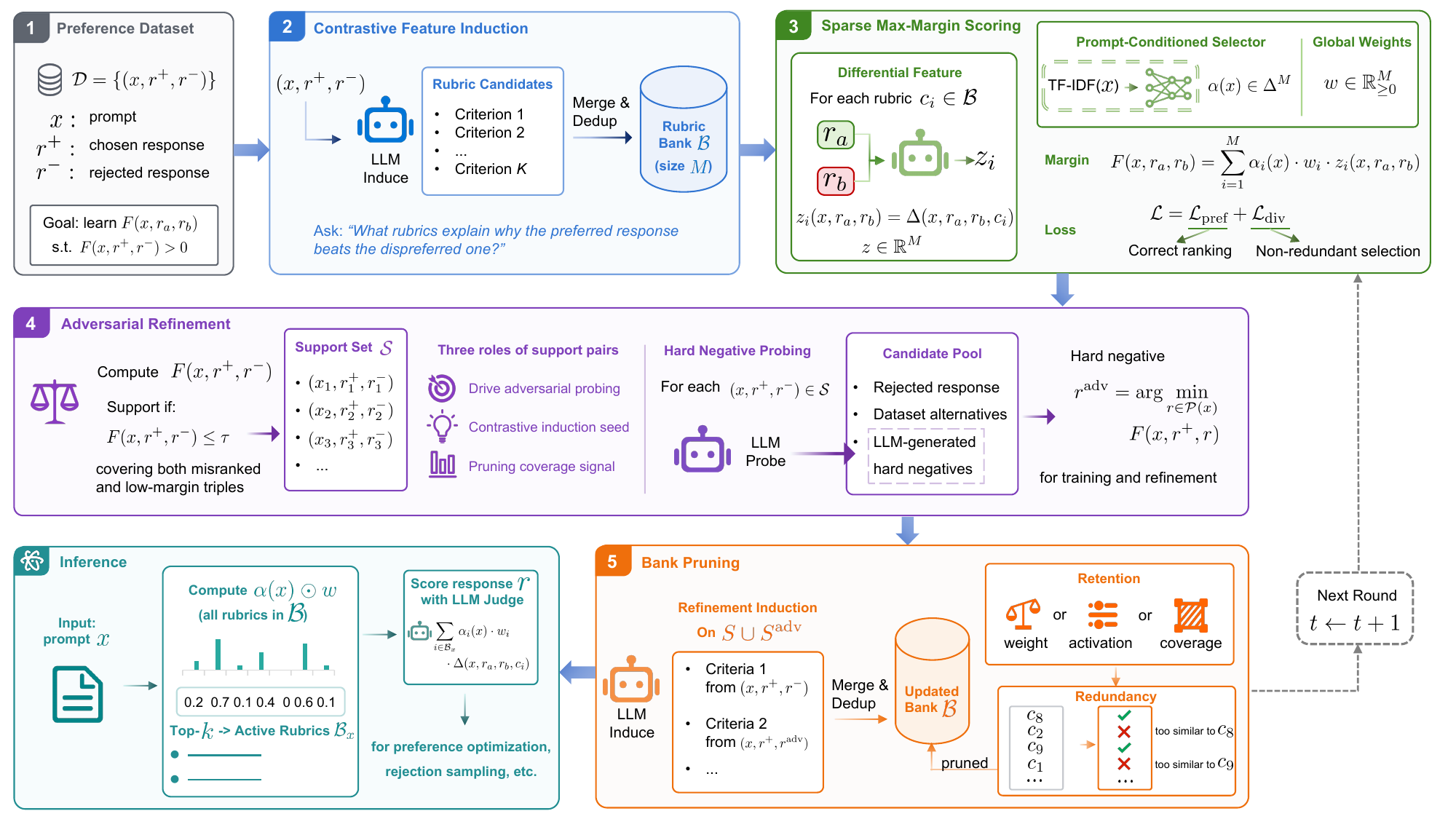}
\caption{\textbf{Overview of \ourmethod{}.} The training loop alternates between fitting $(\alpha, w)$ against a max-margin loss, mining support pairs and adversarial hard negatives, and refining the rubric bank. At inference, the prompt-conditioned selector retrieves top-$k$ support rubrics from the bank to score candidate responses.}
\label{fig:overview}
\end{figure*}

\subsection{Rubric-based Evaluation}

Rubric-based LLM evaluation has attracted growing attention, with recent benchmarks probing rubric authoring by models \citep{zhang2026rubricbench}, rubric-level human-judge agreement \citep{pan2026rubriceval}, instruction following under complex constraints \citep{he2025advancedif}, and hierarchical rubric grading of research replication \citep{starace2025paperbench}.
On the methodological side, OpenRubrics \citep{liu2025openrubrics} synthesises (prompt, rubric) data via contrastive generation, RaR \citep{gunjal2026rubrics} and RIFL \citep{he2025advancedif} use rubrics as RL rewards, RLCER \citep{sheng2026reinforcing} co-trains reasoner and rubricator, and the Implicit-to-Explicit framework of \citet{xie2025auto} fixes a domain-specific core rubric set, requiring re-induction for new domains, and Rubric-ARM \citep{xu2026alternating} alternately trains a rubric proposer and judge via RL for co-evolving rewards.
RRD \citep{shen2026rethinking} mines rubrics from responses sampled by external generators, so the resulting rubrics inherit the coverage and biases of those generators.

\ourmethod{} departs from this landscape in several ways.
Where prior methods \citep{liu2025openrubrics,shen2026rethinking,xie2025auto} optimise rubrics for descriptive coverage, summarising induced rules or sampled responses, we optimise for discriminative margin against support pairs and learn a selector that activates a subset of a dynamic global bank grown with adversarial hard negatives at the boundary.
Unlike RRD~\citep{shen2026rethinking}, our pipeline conditions only on the prompt, and rubrics earn their place by carrying margin on misjudged samples rather than by external sampling.

\subsection{Reward Modeling from Preferences}

Reward models (RMs) translate preference data into scoring functions for alignment, RL fine-tuning, and rejection sampling.
Scalar RMs such as Nemotron-Reward \citep{wang2025helpsteerpreference} and Skywork-Reward \citep{liu2025skywork} dominate RewardBench \citep{lambert2025rewardbench} and RM-Bench \citep{liu2025rm} but produce opaque scores prone to surface bias \citep{sun2026steerrm}.
Generative RMs such as RM-R1 \citep{chen2026rmr}, LMUnit~\citep{saadfalcon2025lmunit}, and critique-out-loud variants~\citep{ankner2024critique} replace the scalar with a free-form rationale, gaining interpretability at the cost of higher inference variance.
Between these, the rubric-based LLM judges above can serve as RMs by aggregating per-rubric judgements into a single preference score post-hoc.

\ourmethod{} sits structurally in this third group and is usable wherever an RM is plugged in, from preference tuning \citep{ziegler2019fine} to RL on open-ended tasks.
Unlike post-hoc rubric aggregation, the pairwise margin learned by \ourmethod{} (Sec.~\ref{sec:margin}) retains scalar-RM geometry \citep{wang2025helpsteerpreference,liu2024skywork,liu2025skywork} while exposing which rubrics fired and how much each contributed.

%% file: sec3-methodology.tex
\section{Methodology}
\label{sec:method}

\subsection{Problem Setup}
\label{sec:setup}

We start from an existing open-source preference dataset $\mathcal{D} = \{(x, r^+, r^-)\}$, where $r^+$ is preferred over $r^-$ for prompt $x$. Our goal is, given $x$, to produce a small set of rubrics that lets a downstream judge score $r^+$ above $r^-$, with particular emphasis on triples near the decision boundary, where self-rubric judges fail \citep{zhang2026rubricbench}. We formalize this through a pairwise margin function $F(x, r_a, r_b)$ that aggregates prompt-conditioned, per-rubric pairwise judgments of $r_a$ against $r_b$, and we require $F(x, r^+, r^-) > 0$.
\ourmethod{} composes $F$ from four components: a global rubric bank as a discriminative feature space (Sec.~\ref{sec:bank}), a sparse max-margin scoring head with a prompt-conditioned selector (Sec.~\ref{sec:margin}), an adversarial boundary-refinement loop driven by support pairs (Sec.~\ref{sec:adv}), and a bank pruning step (Sec.~\ref{sec:prune}).

\subsection{Rubric Bank as Feature Space}
\label{sec:bank}

The criteria that separate chosen from rejected responses are, in practice, drawn from a fairly small and recurring vocabulary, \textit{factuality}, \textit{completeness}, \textit{safety}, \textit{code correctness}, etc., so we expect a finite rubric set to suffice across prompts~\citep{xie2025auto}.
We therefore maintain a global bank $\mathcal{B} = \{c_1, \dots, c_M\}$ of natural-language rubrics, treated as basis dimensions of a discriminative feature space shared across prompts.
$\mathcal{B}$ is grown by \emph{contrastive feature induction}: for each preference pair $(x, r^+, r^-)$, we ask an LLM to propose rubrics that distinguish $r^+$ from $r^-$, then merge candidates into $\mathcal{B}$ via deduplication and similarity collapsing.
Replacing the conventional summarisation framing ``what makes a good response'' \citep{liu2025openrubrics} with a contrastive framing ``what rubric tells these two apart'' surfaces sharper, boundary-relevant criteria rather than generic descriptive ones.

For each rubric $c$, we use an LLM as a pairwise judge that, in a single call, inspects both responses side by side and returns (a) a per-response pass or fail verdict and (b) an explicit better-of-the-two preference.
These signals are combined into a scalar $\Delta(x, r_a, r_b, c) \in [-1.25,\, 1.25]$: the pass/fail difference contributes $\{-1, 0, +1\}$ and the directional preference adds $\pm 0.25$, so positive values favour $r_a$ and negative values favour $r_b$, with magnitude reflecting the strength of distinction.
We collect these per-rubric judgments into the pairwise differential feature

\begin{equation}
z_i(x, r_a, r_b) = \Delta(x, r_a, r_b, c_i),
\label{eq:zfeat}
\end{equation}
so $z \in \mathbb{R}^M$ encodes per rubric which response is favoured and by how much.

\subsection{Sparse Max-Margin Scoring}
\label{sec:margin}

For each prompt, only a small subset of $\mathcal{B}$ is discriminative: a code rubric does little on a safety question.
A prompt-conditioned selector $\alpha(x) \in \Delta^{M}$ identifies this subset, implemented as a small feed-forward network over a TF-IDF \citep{salton1988term} embedding of $x$ followed by a sparsemax projection \citep{martins2016softmax} that drives irrelevant entries exactly to zero (architecture in Appendix~\ref{app:selector}).
This global-bank, local-subspace split lets \ourmethod{} share evidence across prompts while composing the boundary locally per input.

A global learned weight vector $w \in \mathbb{R}_{\ge 0}^M$, shared across prompts, calibrates rubric importance and yields the pairwise margin
\begin{equation}
F(x, r_a, r_b) = \sum_{i=1}^{M} \alpha_i(x) \cdot w_i \cdot z_i(x, r_a, r_b).
\label{eq:margin}
\end{equation}
Geometrically, $w$ acts as the SVM weight vector of a max-margin classifier, $\alpha(x)$ adapts its active subspace per prompt, and the sign of $z_i$ encodes the per-pair direction of each rubric.
Non-negativity of $w$ aligns each rubric with its semantic direction and exposes $w_i$ as an interpretable contribution magnitude that the pruner of Sec.~\ref{sec:prune} thresholds directly.

We fit $(\alpha, w)$ jointly by minimising a composite margin loss
\begin{equation}
\mathcal{L} = \mathcal{L}_{\text{pref}} + \mathcal{L}_{\text{div}}.
\label{eq:loss}
\end{equation}
The preference term $\mathcal{L}_{\text{pref}} = \mathbb{E}_{(x, r^+, r^-)}\!\big[\log\!\big(1 + e^{-F(x, r^+, r^-)}\big)\big]$ enforces a positive margin on each triple.
The diversity term $\mathcal{L}_{\text{div}} = \mathbb{E}_x \!\sum_{i, j} S_{ij}\,\alpha_i(x)\,\alpha_j(x)$, with $S_{ij}$ the textual similarity between rubrics $c_i$ and $c_j$ (form in Appendix~\ref{app:hparams}), penalises the selector for placing weight on near-duplicate rubrics simultaneously.

\subsection{Adversarial Refinement}
\label{sec:adv}

After each fitting round, some triples in $\mathcal{D}$ still have margin $F(x, r^+, r^-) \le \tau$ under the current $(\alpha, w)$, either misranked ($F < 0$) or ranked correctly with too little confidence ($0 \le F \le \tau$).
By analogy with SVM support vectors, we call these \emph{support pairs}.

Support-pair status is recomputed every round, so the same triple may enter and leave the support set as the boundary shifts.
This gives $\mathcal{B}$ a \emph{boundary direction}: it grows toward what is hard for the model right now, not what is hard on average.
Without this re-targeting, induction would accumulate generic descriptive rubrics.

Static pairs saturate once the bank covers obvious $r^+$/$r^-$ distinctions, so we probe the boundary with synthetic hard negatives that expose less-covered failure axes.
For each support pair, an LLM probe receives $(x, r^+)$ and is asked to produce a response that is plausible but inferior to $r^+$ on at least one substantive criterion.
The candidate set $\mathcal{P}(x)$ collects the original $r^-$ as a baseline, any dataset-provided alternatives, and the synthesised candidates, from which we select
\begin{equation}
r^{\text{adv}} = \arg\min_{r \in \mathcal{P}(x)} F(x, r^+, r)
\label{eq:adv}
\end{equation}
under the current $(\alpha, w)$, namely the candidate closest to crossing the boundary.
By construction the preference $r^+ \succ r^{\text{adv}}$ stays well-defined.

Each $r^{\text{adv}}$ enters $\mathcal{L}_{\text{pref}}$ as an additional triple $(x, r^+, r^{\text{adv}})$ averaged with the original $(x, r^+, r^-)$, and seeds the next contrastive-induction round, lifting hard-negative mining \citep{schroff2015facenet,shrivastava2016training} from continuous embeddings to a discrete natural-language feature space.

\subsection{Bank Pruning}
\label{sec:prune}

Support pairs and their adversarial counterparts then drive a boundary-targeted contrastive induction pass, growing new rubrics for failure modes $\mathcal{B}$ does not yet cover.

To keep $\mathcal{B}$ from growing monotonically, each round ends with a four-criterion prune.
An entry is kept if it fires any of three relevance signals, weight $w_i$, activation rate $\mathbb{E}_x[\alpha_i(x)]$, or current-round support-pair coverage, above their respective thresholds. A redundancy filter then sweeps the survivors in descending relevance and drops near-duplicates of higher-ranked entries.
The result is a bounded, non-redundant bank whose surviving rubrics are each high-weight, frequently selected, or boundary-relevant, preserving the interpretability of the trained bank.
The full training procedure is summarised in Algorithm~\ref{alg:svr}.

\definecolor{svrblue}{HTML}{1F6FB2}
\definecolor{svrorange}{HTML}{C2571A}
\definecolor{svrpurple}{HTML}{6B3FA0}
\begin{algorithm}[t]
\small
\caption{\ourmethod{} training procedure}
\label{alg:svr}
\begin{algorithmic}[1]
\Require Dataset $\mathcal{D}$, threshold $\tau$, number of rounds $T$
\Ensure Bank $\mathcal{B}$, parameters $(\alpha, w)$
\State $\mathcal{C} \gets \textcolor{svrblue}{\textsc{ContrastiveInduce}}(\mathcal{D})$ \Comment{initial induction}
\State $\mathcal{S}^{\text{adv}} \gets \emptyset$
\For{$t = 1$ \textbf{to} $T$}
  \State $\mathcal{B} \gets \textsc{BuildBank}(\mathcal{C})$ \Comment{dedup, drop pruned}
  \State $(\alpha, w) \gets \arg\min \mathcal{L}$ on $\mathcal{D} \cup \mathcal{S}^{\text{adv}}$ \Comment{Eq.~\ref{eq:loss}}
  \State $\mathcal{B} \gets \textcolor{svrorange}{\textsc{Prune}}(\mathcal{B}, \alpha, w)$ \Comment{four-criterion}
  \If{$t < T$}
    \State $\mathcal{S} \gets \{(x, r^+, r^-) \in \mathcal{D} \,|\, F(x, r^+, r^-) \le \tau\}$ \Comment{\textcolor{svrpurple}{support pairs}}
    \For{$(x, r^+, r^-) \in \mathcal{S}$}
      \State $r^{\text{adv}}(x) \gets \textcolor{svrpurple}{\textsc{AdversarialProbe}}(x, r^+, \alpha, w)$
    \EndFor
    \State $\mathcal{S}^{\text{adv}} \gets \{(x, r^+, r^{\text{adv}}(x)) \,|\, (x, r^+, r^-) \in \mathcal{S}\}$
    \State $\mathcal{C} \gets \mathcal{C} \cup \textcolor{svrblue}{\textsc{ContrastiveInduce}}(\mathcal{S} \cup \mathcal{S}^{\text{adv}})$ \Comment{boundary re-induction}
  \EndIf
\EndFor
\State \Return $\mathcal{B}, (\alpha, w)$
\end{algorithmic}
\end{algorithm}

After $T$ rounds, $\mathcal{B}$, $\alpha$, and $w$ are frozen.
At inference, \ourmethod{} takes the prompt $x$, computes $\alpha(x) \odot w$ over $\mathcal{B}$, and keeps the top-$k$ entries as the active rubric set $\mathcal{B}_x$, optionally letting the LLM rephrase them to $x$ without changing semantics.
The judge then scores a candidate pair $(r_a, r_b)$ via $F(x, r_a, r_b) = \sum_{i \in \mathcal{B}_x} \alpha_i(x)\,w_i\,\Delta(x, r_a, r_b, c_i)$.
Per-pair inference is one selector pass and $k$ rubric judgments with no test-time rubric generation, leaving \ourmethod{} a lightweight drop-in for downstream RL or rejection-sampling pipelines.

%% file: sec4-experiments.tex
\section{Experiments}
\label{sec:exp}

\subsection{Experimental Setup}

\paragraph{Training data.}
We train \ourmethod{} on HelpSteer3 \citep{NEURIPS2025_3e0271cf}, which provides chosen and rejected responses per prompt.
From $38{,}459$ preference triples spanning general, code, multilingual, and STEM, 13-gram decontamination against our evaluation benchmarks removes \textbf{4.86\%}, leaving $36{,}591$ examples with $5\%$ held out as a dev split.

\paragraph{Training configuration.}
We train for $T = 3$ rounds, with each round fitting $(\alpha, w)$ on the current bank, pruning, and then running adversarial refinement with boundary re-induction to grow the bank for the next round.
All training-time LLM calls use GPT-OSS-120B (OSS) \citep{agarwal2025gpt}, which also serves as the default downstream judge unless otherwise stated.
The selector is a two-layer MLP over TF-IDF prompt features with a sparsemax head, yielding sparse per-prompt rubric weights.
At inference we retain the top-$k = 6$ rubrics, with sensitivity to $k$ analysed in Appendix~\ref{app:topk}.
Additional training details are provided in Appendix~\ref{app:hparams}.

\paragraph{Evaluation benchmarks.}
We evaluate on four benchmarks.
RubricBench \citep{zhang2026rubricbench} is the primary testbed for the discriminative gap, with human-annotated rubrics and hard-to-distinguish pairs whose surface cues often mislead judges.
RewardBench \citep{lambert2025rewardbench}, RewardBench~2 \citep{malik2025rewardbench}, and RM-Bench \citep{liu2025rm} are general reward-modelling benchmarks that test pairwise agreement with human preferences across diverse subtasks.
We report pairwise accuracy throughout.

\paragraph{Baselines.}
We compare against five baselines.
\textbf{Reward models.} We include the state-of-the-art scalar RM \texttt{Skywork-Reward-V2-Llama-3.1-8B} (Skywork) \citep{liu2025skywork} and the generative RM \texttt{RM-R1-Qwen2.5-Instruct-7B} (RM-R1) \citep{chen2026rmr}.
\textbf{Vanilla LLM-as-a-Judge.} The judge scores directly from $(x, r_a, r_b)$.
\textbf{LLM with self-generated rubrics} (Self-Rubrics). The judge LLM writes its own rubric per prompt before scoring. \citep{liu2023g}
\textbf{Rubric-ARM} \citep{xu2026alternating}. The authors release two dedicated models, one for proposing rubrics and one for judging with them.
\textbf{RRD} \citep{shen2026rethinking}. This method mines rubrics from preference responses sampled by external generators, which we reproduce following the original protocol (details in Appendix~\ref{app:rrd_repro}).

\input{tables/rubricbench}

\subsection{RubricBench Results}

Table~\ref{tab:rubricbench} shows that \ourmethod{} reaches \textbf{82.8} overall, outperforming every non-oracle baseline by a large margin.
Against methods without explicit rubrics, \ourmethod{} improves over Skywork and the strongest direct LLM judge by more than \textbf{21} points, showing that RubricBench is not solvable by implicit signals alone, whether from a scalar head or from free-form judge reasoning.
The same conclusion holds for rubric-based baselines: with the same OSS judge, \ourmethod{} is \textbf{23.8} points above self-generated rubrics, and it is also far ahead of Rubric-ARM and RRD.
These comparisons suggest that the main advantage of \ourmethod{} is the quality of the selected rubrics: they are learned to discriminate hard response pairs, rather than being generic criteria produced without any discrimination signal.

Compared with the human-annotated oracle, \ourmethod{} narrows the gap from OSS self-generated rubrics from $24.1$ points to only $0.3$.
The domain-level results show where this gap is closed: \ourmethod{} slightly exceeds the oracle on Chat, matches it on IF, and remains close on STEM and Code.
The main remaining difference is Safety ($83.8$ vs. $92.5$), where human rubrics still better capture rare, high-stakes distinctions such as refusal boundaries and safety caveats.
Since the oracle requires per-instance human rubric writing, these results show that \ourmethod{} provides a much more scalable route to near-human rubric performance.

\subsection{Reward-Modelling Benchmarks}

\input{tables/rmbenches}

Rubric-based evaluation is closely related to reward modelling, so we further evaluate on three general reward-modelling benchmarks to test broader applicability beyond the rubric-focused setting.
As shown in Table~\ref{tab:rmbenches}, Skywork achieves the highest average score ($90.3$), as it is a scalar RM purpose-trained specifically for reward modelling. \ourmethod{} (\textbf{87.2}) edges past the vanilla OSS baseline ($87.0$), since holistic heuristics already account for most of the signal on these broad preference benchmarks and leave limited headroom for structured rubrics to push the score further.
Indeed, both Self-Rubrics ($78.5$) and Rubric-ARM ($68.1$) fall well below OSS, as prompt-only and pipeline-generated criteria overwrite the holistic prior without installing a sharper substitute. \ourmethod{}, by contrast, \emph{preserves} that prior, further indicating that our learned rubrics are more precise than those of the baselines.

At the domain level, \ourmethod{} reaches $81.9$ on Precise IF (RewardBench~2), outperforming Skywork ($64.1$) by \textbf{17.8} points, extending the IF advantage from RubricBench.
On RM-Bench, \ourmethod{} stays within \textbf{1.1} points of Skywork across all three difficulty tiers, showing consistent competitiveness regardless of pair difficulty.
Safety, however, remains the consistent weak spot (\textbf{82.8}/\textbf{72.4} on RewardBench/RewardBench~2 vs. Skywork at \textbf{95.7}/\textbf{96.7}), a domain on which Skywork has been specifically trained with a dedicated preference dataset.
\ourmethod{} thus complements alignment-tuned RMs on instruction-heavy tasks while conceding the safety advantage to models purpose-built for it.

%% file: tables/rubricbench.tex
\begin{table*}[t]
\centering
\small
\setlength{\tabcolsep}{4pt}
\renewcommand{\arraystretch}{1.05}
\caption{Pairwise accuracy (\%) on RubricBench broken down by domain. \textbf{Bold} marks the best per column, \underline{underline} the second best. For Rubric-ARM we list backbones as \emph{proposer / judge} when they differ.}
\label{tab:rubricbench}
\begin{tabular}{p{4cm}lcccccc}
\toprule
\multirow{2}{*}{\textbf{Method}} & \multirow{2}{*}{\textbf{Model}} & \multicolumn{5}{c}{\textbf{Domain Accuracy}} & \textbf{Overall} \\
\cmidrule(lr){3-7}
 & & Chat & IF & STEM & Code & Safety & \textbf{Acc} \\
\midrule
\rowcolor{gray!10}
\multicolumn{8}{l}{\textit{\textbf{Human-Annotated Oracle}}} \\
Oracle Rubrics                         & OSS            & \underline{80.6} & \textbf{85.5} & \textbf{83.6} & \textbf{86.4} & \textbf{92.5} & \textbf{83.1} \\
\rowcolor{gray!10}
\multicolumn{8}{l}{\textit{\textbf{Reward Models}}} \\
Scalar RM   & Skywork            & 63.0 & 69.4 & 58.0 & 52.0 & 77.5 & 61.0 \\
Generative RM   & RM-R1     & 43.1 & 39.5 & 50.4 & 47.2 & 53.8 & 46.0 \\
\rowcolor{gray!10}
\multicolumn{8}{l}{\textit{\textbf{Vanilla LLM-as-a-Judge}}} \\
Direct pairwise judge                  & OSS            & 44.6 & 69.4 & 60.8 & 58.7 & 60.0 & 55.2 \\
Direct pairwise judge                  & Gemini-3-Flash            & 50.0 & 69.4 & 66.8 & 63.1 & 36.3 & 57.9 \\
Direct pairwise judge                  & GPT-5.4            & 53.1 & 66.1 & 69.6 & 64.9 & 56.3 & 61.1 \\
\rowcolor{gray!10}
\multicolumn{8}{l}{\textit{\textbf{Dedicated Rubric Framework}}} \\
Rubric-ARM                             & RubricARM-8B                           & 51.0 & 43.6 & 51.6 & 50.2 & 28.8 & 48.6 \\
Rubric-ARM                             & RubricARM-8B / OSS             & 48.3 & 58.1 & 61.2 & 54.2 & 25.0 & 52.0 \\
\rowcolor{gray!10}
\multicolumn{8}{l}{\textit{\textbf{Rubric-Augmented LLM-as-a-Judge}}} \\
Self-Generated Rubrics                 & OSS    & 54.7 & 69.4 & 64.4 & 59.0 & 48.8 & 59.0          \\
Self-Generated Rubrics                 & Gemini-3-Flash   & 55.2 & 70.2 & 65.6 & 60.9 & 50.0 & 60.1         \\
Self-Generated Rubrics                 & GPT-5.4 & 55.9 & \underline{71.0} & 65.6 & 60.1 & 51.2 & 60.3           \\
RRD         & OSS            & 65.3 & 64.0 & 56.8 & 53.8 & 52.8 & 57.6 \\
\rowcolor{cyan!10}
\textbf{\ourmethod{} (Ours)}           & OSS            & \textbf{81.8} & \textbf{85.5} & \underline{82.4} & \underline{83.4} & \underline{83.8} & \underline{82.8} \\

\bottomrule
\end{tabular}
\end{table*}

%% file: tables/rmbenches.tex
\begin{table*}[t]
\centering
\footnotesize
\setlength{\tabcolsep}{4pt}
\renewcommand{\arraystretch}{1.1}
\caption{Pairwise accuracy (\%) on RewardBench, RewardBench 2, and RM-Bench. We report all subdomain scores. The final column (Avg.) reports the mean accuracy across all preceding columns. Self-Rubrics denotes the self-generated rubrics baseline. \textbf{\textbf{Bold}} marks the best per column, \underline{underline} the second best.}
\label{tab:rmbenches}
\resizebox{\textwidth}{!}{%
\begin{tabular}{lcccc|ccccc|ccc|c}
\toprule
\multirow{2}{*}{\textbf{Method}} & \multicolumn{4}{c|}{\textbf{RewardBench}} & \multicolumn{5}{c|}{\textbf{RewardBench 2}} & \multicolumn{3}{c|}{\textbf{RM-Bench}} & \multirow{2}{*}{\textbf{Avg.}} \\
\cmidrule(lr){2-5} \cmidrule(lr){6-10} \cmidrule(lr){11-13}
 & Chat & Chat Hard & Reasoning & Safety & Factuality & Precise IF & Math & Safety & Focus & Easy & Normal & Hard & \\
\midrule
Skywork & \textbf{99.4} & \textbf{90.4} & \textbf{99.6} & \textbf{95.7} & \textbf{84.5} & 64.1 & 77.6 & \textbf{96.7} & \textbf{98.4} & \textbf{97.3} & \textbf{95.4} & \textbf{84.9} & \textbf{90.3} \\
RM-R1 & 95.8 & 69.7 & 83.2 & 83.2 & 44.9 & 29.5 & 72.6 & 60.7 & 84.4 & 82.6 & 72.9 & 55.4 & 69.6   \\
OSS & \underline{96.1} & \underline{88.8} & 98.3 & \underline{91.4} & 76.6 & 67.5 & \textbf{85.9} & 83.3 & 89.3 & 93.9 & 91.1 & 81.8 & 87.0 \\
Rubric-ARM & 91.1 & 82.2 & 81.1 & 58.4 & 61.7 & 53.1 & 72.7 & 40.0 & \underline{89.9} & 72.6 & 64.6 & 50.2 & 68.1 \\
Self-Rubrics OSS & 80.7 & 72.9 & 87.0 & 88.1 & 73.2 & \underline{70.2} & \underline{85.4} & \underline{84.4} & 80.4 & 79.2 & 72.5 & 68.1 & 78.5 \\
\midrule
\textbf{\ourmethod{} (ours)} & 95.5 & 87.3 & \underline{98.4} & 82.8 & \underline{80.6} & \textbf{81.9} & 85.3 & 72.4 & 87.5 & \underline{96.2} & \underline{94.6} & \underline{83.9} & \underline{87.2} \\
\bottomrule
\end{tabular}%
}
\end{table*}

%% file: sec5-analysis.tex
\section{Analysis}
\label{sec:analysis}

This section analyses where the gains of \ourmethod{} come from, how its training process converges, how stable the learned rubrics are across judges, and how its decisions compare with human-annotated rubrics.

\subsection{Ablation Studies}
\label{sec:analysis_ablation}

\input{tables/ablation}

Table~\ref{tab:ablation} isolates each component in \ourmethod{}, removing one at a time under the same judge and protocol as the main experiments.

Each row disables one design from Sec.~\ref{sec:method}. ``$-$ contrastive induction'' swaps the discriminative rubric prompt for the conventional ``what makes a good response''. ``$-$ support-pair mining'' drops the margin threshold $\tau$ and uses every pair for re-induction. ``$-$ adversarial probe'' skips the synthesised $r^{\text{adv}}$. ``$-$ diversity term $\mathcal{L}_{\text{div}}$'' drops the redundancy penalty.

The last two rows are inference-time variants. ``$\to$ uniform top-$k$ weighting'' keeps top-$k$ selection by $\alpha(x) \odot w$ but assigns each selected rubric weight $1$ in the margin sum, isolating the learned weighting from the learned selection. ``$\to$ pointwise inference'' swaps the pairwise judge for a pointwise scorer, $z_i = s(x, r_a, c_i) - s(x, r_b, c_i)$. The modest drop to $79.7$ shows the learned rubrics remain usable as a single-response scoring head for downstream uses such as RL reward modelling.

\begin{figure}[t]
\centering
\includegraphics[width=\linewidth]{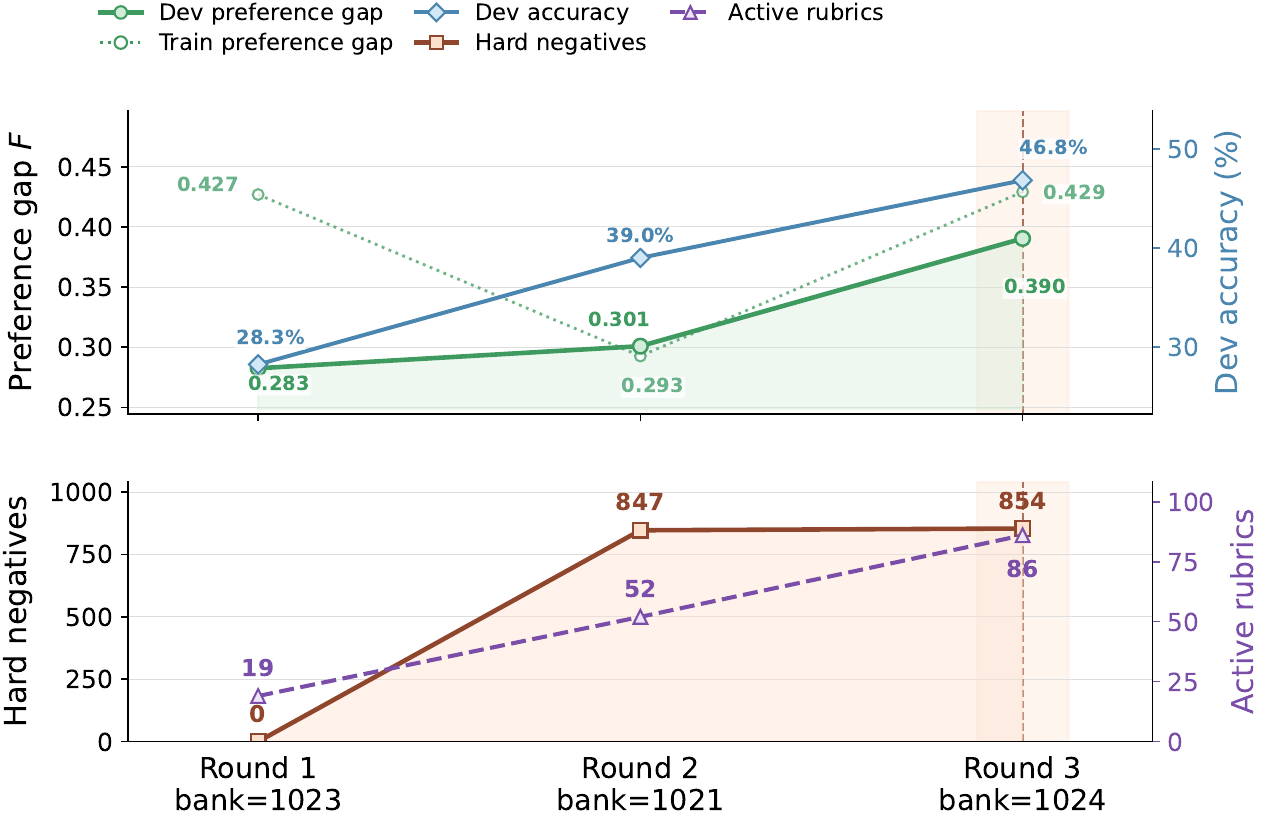}
\caption{Training dynamics of \ourmethod{} across three refinement rounds. The development margin and accuracy rise monotonically, and the train-dev margin gap closes from $0.144$ to $0.039$ by round three, marking the boundary at which we stop refinement.}
\label{fig:training_dynamics}
\end{figure}

\begin{figure*}[t]
\centering
\begin{minipage}[t]{0.48\textwidth}
\begin{databox}{Case 1. \ourmethod{}-only correct}
\small
\textbf{Prompt} (general). Compose a polite email reply explaining the user has no partners yet and will follow up later.\\[2pt]
\textbf{Resp A} ($244$ ch). Direct one-paragraph reply that mirrors the requested message.\\
\textbf{Resp B} ($482$ ch). Multi-paragraph reply that volunteers unrequested topics such as language practice sessions.\\[2pt]
\textbf{\ourmethod{}}. Rubrics on \emph{stick to requested content} and \emph{avoid unrequested expansion} favour A. Margin $+6$. (correct)\\
\textbf{Oracle}. Rubrics on \emph{information completeness} favour B. (wrong)\\
\textbf{Chosen}: A.
\end{databox}
\end{minipage}\hfill
\begin{minipage}[t]{0.48\textwidth}
\begin{databox}{Case 2. Oracle-only correct}
\small
\textbf{Prompt} (human-preference). Polish my Chinese graduate-school self-introduction, with $800$-character original.\\[2pt]
\textbf{Resp A}. Four bullet points of suggestions for how to improve the original.\\
\textbf{Resp B}. The full revised self-introduction text.\\[2pt]
\textbf{\ourmethod{}}. Rubrics on \emph{clear bullet structure} and \emph{coherence} favour A. Margin $-1$. (wrong)\\
\textbf{Oracle}. Rubric ``returns a revised self-introduction, not an analysis'' favours B. (correct)\\
\textbf{Chosen}: B.
\end{databox}
\end{minipage}
\caption{Two RubricBench cases where \ourmethod{} and the human-rubric oracle disagree. Case~1 exemplifies a tendency of the oracle to reward coverage at the expense of constraint adherence. Case~2 exemplifies the absence in \ourmethod{} of a \emph{did-the-response-perform-the-task} rubric.}
\label{fig:case_complementary}
\end{figure*}

\subsection{Training Dynamics}
\label{sec:analysis_dynamics}

Figure~\ref{fig:training_dynamics} tracks $F(x, r^+, r^-)$ on the train and dev splits, dev accuracy, hard negatives, and active rubrics. Dev margin and accuracy rise monotonically, confirming refinement widens the gap on unseen pairs. The V-shape in the train margin ($0.427{\to}0.293{\to}0.429$) is the expected signature of boundary mining: each round injects unseparated pairs, the average drops, and the re-fit bank recovers. The decisive stopping signal is that the train-dev divergence closes from $0.144$ to $0.039$, leaving no generalisation gap for a fourth round to close. At its stable scale of roughly $850$, the hard-negative pool adds only seven pairs beyond the $847$ from round two, while active rubrics rise from $19$ to $52$ to $86$ on a bank held near $1023$, indicating reorganisation rather than expansion.

\subsection{Transfer of Rubrics Across Judges}
\label{sec:analysis_models}

\input{tables/model_transfer}

A rubric bank is only useful if it generalises beyond the judge it was trained against. We therefore freeze the bank trained with GPT-OSS-120B \citep{agarwal2025gpt} and swap in four additional judges that span scale and family: GPT-OSS-20B \citep{agarwal2025gpt}, GPT-4o-mini \citep{hurst2024gpt}, DeepSeek-V4-Flash \citep{deepseekai2026deepseekv4}, and Qwen3.5-397B-A17B \citep{qwen35blog}.

As shown in Table~\ref{tab:model_transfer}, \ourmethod{} is the best rubric source for every one of the five judges, averaging \textbf{80.6} versus $57.4$ for Vanilla and $57.3$ for Self-Rubrics, a gap of more than \textbf{23} points over both baselines. Self-Rubrics oscillates around Vanilla without a consistent direction, indicating that prompt-only criteria offer no reliable lift on top of direct judging. \ourmethod{}, in contrast, retains a large margin whether the judge is a smaller model or drawn from a different family, showing that the rubric bank captures discrimination signals that are largely judge-agnostic. In practice, the bank can be trained once and reused across capable judges without retraining.

\subsection{Complementarity with Human Rubrics}
\label{sec:analysis_residual}

\begin{table}[t]
\centering
\small
\setlength{\tabcolsep}{6pt}
\caption{Per-example correctness overlap between \ourmethod{} and the human-rubric oracle on RubricBench.}
\label{tab:error_overlap}
\begin{tabular}{lrr}
\toprule
Outcome & Count & \% \\
\midrule
Both correct                & $798$   & $69.6$ \\
\ourmethod{}-only correct   & $152$   & $13.3$ \\
Oracle-only correct         & $155$   & $13.5$ \\
Both wrong                  & $42$    & $3.7$  \\
\bottomrule
\end{tabular}
\end{table}

\ourmethod{} nearly matches the human-rubric oracle on RubricBench, reaching \textbf{82.8} compared with \textbf{83.1}.
Table~\ref{tab:error_overlap} crosses per-example correctness on the $1{,}147$ pairs and shows the two sources disagree on $26.8\%$ of them, with each uniquely correct on roughly $13\%$.
\ourmethod{} thus reaches oracle accuracy through a different set of decisions, capturing an independent judgment signal.

Two near-balanced cases in Figure~\ref{fig:case_complementary} illustrate the disagreement. The oracle, with hand-written rubrics sometimes overweights surface specifics or rewards coverage at the expense of the requested form (Case~1). The \ourmethod{} bank, biased toward presentation and quality dimensions, lacks rubrics for whether the response actually performed the requested task (Case~2). App.~\ref{app:residual} expands these failure modes with domain-level net advantages and a taxonomy of the $42$ shared failures into subjective preference, capability ceiling, and label noise.

%% file: tables/ablation.tex
\begin{table}[t]
\centering
\small
\setlength{\tabcolsep}{5pt}
\caption{Ablation on RubricBench. $\Delta$ is relative to \ourmethod{}.}
\label{tab:ablation}
\begin{tabular}{lcc}
\toprule
Variant & Accuracy & $\Delta$ \\
\midrule
\ourmethod{} & \textbf{82.8} & -- \\
\quad $-$ contrastive induction & 73.5 & $-$9.3 \\
\quad $-$ support-pair mining & 77.6 & $-$5.2 \\
\quad $-$ adversarial probe & 79.0 & $-$3.8 \\
\quad $-$ diversity term $\mathcal{L}_{\text{div}}$ & 81.1 & $-$1.7 \\
\midrule
\quad $\to$ uniform top-$k$ weighting & 64.3 & $-$18.5 \\
\quad $\to$ pointwise inference & 79.7 & $-$3.1 \\
\bottomrule
\end{tabular}
\end{table}

%% file: tables/model_transfer.tex
\begin{table}[t]
\centering
\small
\setlength{\tabcolsep}{4pt}
\caption{RubricBench accuracy across five judge backbones. The \ourmethod{} bank is trained once with GPT-OSS-120B and kept fixed at inference. \textbf{Bold} marks the best method per row, \underline{underline} the second best.}
\label{tab:model_transfer}
\begin{tabular}{lccc}
\toprule
Judge & Vanilla & Self-Rubrics & \ourmethod{} \\
\midrule
GPT-OSS-120B & 55.2 & \underline{59.0} & \textbf{82.8} \\
GPT-OSS-20B & \underline{55.3} & 53.9 & \textbf{77.4} \\
GPT-4o-mini & \underline{56.1} & 51.8 & \textbf{80.3} \\
DeepSeek-V4-Flash & 56.5 & \underline{64.4} & \textbf{80.2} \\
Qwen3.5-397B-A17B & \underline{63.7} & 57.6 & \textbf{82.4} \\
\midrule
Average & \underline{57.4} & 57.3 & \textbf{80.6} \\
\bottomrule
\end{tabular}
\end{table}

%% file: sec5-conclusion.tex
\section{Conclusion}
\label{sec:conclusion}

\ourmethod{} casts rubric construction as boundary learning over preference data. The framework jointly mines, weights, and refines a global rubric bank, and at inference retrieves only the rubrics most relevant to each prompt.
On RubricBench, the resulting bank approaches human-written rubrics, and on RewardBench~1\&2 and RM-Bench it remains competitive with dedicated reward models.
Together, these results suggest that the discriminative gap in current rubric-based judging is largely an artefact of how rubrics are obtained, and that boundary-defining criteria can be induced from preference signals alone without per-instance human authoring.

%% file: appendix.tex
\appendix

\section{Hyperparameters and Implementation Details}
\label{app:hparams}

Table~\ref{tab:hparams} lists the single hyperparameter configuration used across all experiments and ablations in this paper.

\input{tables/hparams}

The global weight vector $w$ in Eq.~\ref{eq:margin} is constrained to be non-negative.
We enforce this by reparameterising $w = \mathrm{softplus}(\tilde w)$ with an unconstrained $\tilde w \in \mathbb{R}^M$, and optimising $\tilde w$ directly.
This keeps the constraint exact under standard gradient updates without projection or clipping.

The similarity matrix $S$ used in the diversity term of Eq.~\ref{eq:loss} and in the redundancy filter of Sec.~\ref{sec:prune} is defined directly on rubric texts.
For each pair of rubrics $(c_i, c_j)$, we compute a base similarity $\mathrm{sim}(c_i, c_j) = \max\!\big(J(c_i, c_j),\, R(c_i, c_j)\big)$, where $J$ is the Jaccard overlap of lower-cased content tokens (stopwords and single-character tokens removed) and $R$ is the \texttt{difflib.SequenceMatcher} ratio over alphanumeric-normalised text.
We then threshold and rescale, $S_{ij} = \max\!\big(0,\, \mathrm{sim}(c_i, c_j) - \tau_{\mathrm{red}}\big) / (1 - \tau_{\mathrm{red}})$ for $i \ne j$ with $S_{ii} = 0$, so only pairs above the redundancy threshold $\tau_{\mathrm{red}}$ contribute.
$S$ is recomputed once per round after $\mathcal{B}$ is updated and held fixed during the subsequent fitting step.

\section{Prompt-Conditioned Selector}
\label{app:selector}

\paragraph{Architecture.}
The prompt-conditioned selector $\alpha(x) \in \Delta^{M}$ maps a TF-IDF embedding of the prompt to a sparse distribution over the rubric bank.
The vectoriser is a \texttt{scikit-learn} \texttt{TfidfVectorizer} \citep{pedregosa2011scikit} with unicode-normalised character folding, $n$-gram range $(1, 2)$, vocabulary capped at $4{,}096$ features, and minimum document frequency $1$.
Writing $\phi(x) \in \mathbb{R}^{d}$ for the resulting TF-IDF feature vector with $d \le 4{,}096$, the forward pass is
\begin{equation}
\alpha(x) = \mathrm{sparsemax}\!\big(W_2\, \mathrm{ReLU}(W_1 \phi(x) + b_1) + b_2\big),
\label{eq:selector}
\end{equation}
with a single hidden layer of dimension $256$, so $W_1 \in \mathbb{R}^{256 \times d}$ and $W_2 \in \mathbb{R}^{M \times 256}$.
The $\mathrm{sparsemax}$ projection \citep{martins2016softmax} maps logits onto the probability simplex while driving sub-threshold entries exactly to zero, yielding a small active rubric set per prompt without an explicit top-$k$ truncation at training time.
The per-prompt rubric weighting consumed by the max-margin scoring head of Eq.~\ref{eq:margin} is $\alpha(x) \odot w$, where the global weights $w$ are constrained via the softplus reparameterisation \citep{dugas2000incorporating} described in Appendix~\ref{app:hparams}.

\paragraph{Why a lightweight selector suffices.}
Each rubric $c_i \in \mathcal{B}$ is mined from a specific preference pair $(x_i, r^+_i, r^-_i)$ in Sec.~\ref{sec:bank}, so its discriminative scope is inherited from the surface characteristics of $x_i$. Let $\tau(x)$ denote a latent prompt-type indicator (e.g., coding, refusal, math) and write $\tau_i \equiv \tau(x_i)$. The bank partitions into clusters $\mathcal{C}_t = \{i : \tau_i = t\}$, and the ideal routing concentrates mass on the cluster matching the test prompt,
\begin{equation}
\alpha^{\star}_i(x) \;\propto\; \mathbf{1}\!\big[\tau_i = \tau(x)\big]\, \cdot\, d_i,
\label{eq:ideal-routing}
\end{equation}
with $d_i \ge 0$ encoding how discriminative $c_i$ was on its source pair. Recovering $\alpha^{\star}$ reduces to multi-label classification of $x$ over a small number of latent types, mapped onto $M$ rubric slots through the partition. Coarse lexical features suffice because $\tau(x)$ is largely carried by keywords and short phrases (``python'', ``write a function'', ``is it safe''), well within the capacity of a two-layer MLP over TF-IDF inputs.

The max-margin objective (Eq.~\ref{eq:loss}) drives the learned $\alpha$ toward $\alpha^{\star}$ end to end. Differentiating $\mathcal{L}_{\text{pref}}$ through Eq.~\ref{eq:margin} gives
\begin{equation}
\frac{\partial \mathcal{L}_{\text{pref}}}{\partial \alpha_i(x)} \;=\; -\,\sigma\!\big(-F(x, r^+, r^-)\big)\, \cdot\, w_i\, z_i(x, r^+, r^-),
\label{eq:routing-grad}
\end{equation}
with $\sigma$ the logistic and $w_i \ge 0$. The sign of $z_i$ sets the routing direction. Rubrics with $z_i > 0$ on the supervising pair receive descent pressure that raises their pre-sparsemax logits, while rubrics with $z_i < 0$ are pushed below the simplex boundary, at which point sparsemax projects them onto exactly zero. Routing errors therefore self-correct from the same supervision signal that fits $w$, without any auxiliary router loss.

The selector serves only as a retrieval layer that narrows the candidate set, and the per-rubric judgment is performed by the full LLM judge. The top-$k$ sensitivity analysis in Appendix~\ref{app:topk} supports this split, with accuracy already reaching $81.52$ at $k = 1$.

\section{Prompt Templates}
\label{app:prompts}

\ourmethod{} invokes the LLM in three roles, each with a dedicated prompt template.
We give condensed forms below.
Full templates with formatting instructions are released with the code.

\paragraph{Contrastive feature induction.}
Given a preference pair, the LLM proposes rubrics that distinguish $r^+$ from $r^-$.

\begin{promptbox}{Contrastive Feature Induction}
\small\ttfamily
You are given a prompt and two responses, one preferred (A) and one not (B).\\[2pt]
Propose 2 to 6 specific, verifiable rubrics such that A satisfies each rubric strictly more than B.\\[2pt]
Each rubric must be a single sentence describing a discriminative criterion, not a generic quality dimension.\\[2pt]
\\
Prompt: \{x\}\\
Response A (preferred): \{r\_plus\}\\
Response B (rejected): \{r\_minus\}\\
\\
Return only JSON in the following schema:\\
\{\\
"contrastive\_rubrics": [\\
~~~~\{\\
~~~~~~"facet": "correctness | format | coverage | grounding | tool\_use | coherence | safety | style | language | conciseness | reasoning",\\
~~~~~~"importance": "critical | major | minor",\\
~~~~~~"rubric": "<one reusable rubric>",\\
~~~~~~"grounding": "<why this rubric explains the preference gap>"\\
~~~~\}\\
~~]\\
\}
\end{promptbox}

\paragraph{Adversarial probe.}
For each support pair, the LLM synthesises a candidate that is plausible but inferior to $r^+$.

\begin{promptbox}{Adversarial Probe}
\small\ttfamily
You are given a prompt and a high-quality response.\\[2pt]
Write an alternative response that looks plausible to a casual reader but is inferior to the original on at least one substantive criterion.\\[2pt]
Do not signal the inferiority explicitly.\\[2pt]
\\
Prompt: \{x\}\\
Reference response: \{r\_plus\}\\
\\
Output only the alternative response.
\end{promptbox}

\paragraph{Pairwise rubric judge.}
For each rubric, the judge inspects both candidates side by side in a single call and returns a per-rubric pass or fail verdict for each candidate plus a directed preference, which we combine into the scalar $\Delta(x, r_a, r_b, c)$ used as $z_i$ in Eq.~\ref{eq:zfeat}.

\begin{promptbox}{Pairwise Rubric Judge}
\small\ttfamily
Given a prompt, two candidate responses A and B, and a list of rubrics, judge each rubric pairwise.\\[2pt]
For every rubric, output: (i) a pass or fail verdict for candidate A, (ii) a pass or fail verdict for candidate B, and (iii) which candidate is better on that rubric (A or B).\\[2pt]
\\
Prompt: \{x\}\\
Candidate A: \{r\_a\}\\
Candidate B: \{r\_b\}\\
Rubrics: \{c\_1, ..., c\_k\}\\
\\
Output a JSON object with a \texttt{rubric\_comparisons} list, using one comparison per rubric in the given order.
\end{promptbox}

\paragraph{Vanilla pairwise judge (baseline).}
For the Vanilla LLM-as-a-Judge row in Table~\ref{tab:rubricbench}, the same judge model scores directly from $(x, r_a, r_b)$ without any rubric, support pair, or reference label. The \texttt{rubrics} and \texttt{reference\_rubrics} fields of each RubricBench example are stripped before prompt construction so the judge sees only the instruction and the two candidate responses. The final verdict is extracted by matching the last \texttt{[[A]]} or \texttt{[[B]]} token in the model output.

\begin{promptbox}{Vanilla LLM-as-a-Judge}
\small\ttfamily
Please act as an impartial judge and evaluate the quality of two candidate responses to the user question displayed below.\\[2pt]
\\
You should choose the candidate response that follows the user's instructions and answers the user's question better. Your evaluation should consider as many factors as possible. Begin your evaluation by comparing the two candidate responses and provide thorough reasoning. Avoid any position biases and ensure that the order in which the responses were presented does not influence your decision. Do not allow the length of the responses to influence your evaluation. Do not favor either candidate label. Be as objective as possible.\\[2pt]
\\
After providing your reasoning, output your final verdict by strictly following this format: \mbox{[[A]]} if candidate response A is better, \mbox{[[B]]} if candidate response B is better.\\[2pt]
\\
{[Instruction]}\\
\{x\}\\
\\
{[The Start of Candidate Response A]}\\
\{r\_a\}\\
{[The End of Candidate Response A]}\\
\\
{[The Start of Candidate Response B]}\\
\{r\_b\}\\
{[The End of Candidate Response B]}
\end{promptbox}

\paragraph{Rubric-Augmented LLM-as-a-Judge (baseline).}
For the Self-Generated Rubrics rows in Table~\ref{tab:rubricbench}, the same judge model first writes its own rubrics from the prompt alone and then scores the two candidates against those rubrics. No support pair, training signal, or reference label is used, and the \texttt{rubrics} and \texttt{reference\_rubrics} fields of each RubricBench example are stripped before prompt construction. The pipeline has two single-turn stages, rubric generation from $x$ and pairwise rubric scoring of $(r_a, r_b)$, both restricted to a binary A or B verdict at the response level.

\textbf{Rubric generation.} A single call reads only the user prompt and emits a short list of atomic, response-facing rubrics together with a one-sentence note that records the expected answer shape.

\begin{promptbox}{Self-Generated Rubrics: Rubric Generation}
\small\ttfamily
You are constructing prompt-only evaluation rubrics for a single user turn.\\[2pt]
Use only the prompt below. Do not answer it. Do not invent details about any candidate response.\\[2pt]
\\
Internally follow three steps and then output only the final rubric list:\\[2pt]
(1) Task parse. Identify the task type, the required output shape, and any hard or soft constraints that are explicitly stated in the prompt.\\[2pt]
(2) Answer sketch. Outline at a high level what a strong answer needs to cover and list the most likely failure modes.\\[2pt]
(3) Rubric synthesis. Write 4 to 7 atomic rubrics, each phrased as one independently checkable question about the response.\\[2pt]
\\
Rules:\\[2pt]
Each rubric covers exactly one criterion and is self-contained.\\[2pt]
Each rubric cites either an explicit prompt requirement, an inferred task requirement, or a likely failure mode.\\[2pt]
Do not reward verbosity, formatting polish, or stylistic flourish unless the prompt explicitly requires them.\\[2pt]
For objective tasks, do not reveal a guessed answer key.\\[2pt]
\\
Prompt: \{x\}\\
\\
Output a JSON object with two fields: \texttt{rubrics}, a list of rubric strings, and \texttt{rubric\_note}, a single sentence about the expected answer shape.
\end{promptbox}

\textbf{Pairwise rubric scoring.} A single call evaluates both candidates against the generated rubrics and returns a per-rubric pass or fail verdict for each candidate, a per-rubric preferred side, and an overall preferred candidate. The verdict space is restricted to A or B at every level, with no tie option, so the prediction can be mapped back to chosen or rejected without ambiguity.

\begin{promptbox}{Self-Generated Rubrics: Pairwise Scoring}
\small\ttfamily
You are an impartial judge. Evaluate two candidate responses against the rubrics below.\\[2pt]
Use only the rubrics as the evaluation checklist. Do not introduce extra criteria. Do not let response length, formatting polish, or writing style influence your judgment unless a rubric explicitly asks for them. Avoid position bias.\\[2pt]
\\
For every rubric, decide:\\[2pt]
(i) a pass or fail verdict for Candidate A,\\[2pt]
(ii) a pass or fail verdict for Candidate B,\\[2pt]
(iii) which candidate is better on that rubric, A or B.\\[2pt]
\\
After completing all rubric-level judgments, output a single overall preferred candidate, A or B.\\[2pt]
\\
Prompt: \{x\}\\
Rubrics: \{c\_1, ..., c\_k\}\\
Rubric note: \{note\}\\
Candidate A: \{r\_a\}\\
Candidate B: \{r\_b\}\\
\\
Output a JSON object with two fields: \texttt{rubric\_comparisons}, a list of items with keys \texttt{rubric\_id}, \texttt{candidate\_a\_verdict}, \texttt{candidate\_b\_verdict}, \texttt{better}, and \texttt{reason}, and \texttt{preferred\_candidate}, the overall winner A or B.
\end{promptbox}

\section{LLM Inference Configuration}
\label{app:inference}

All LLM calls in this paper share the same model and decoding configuration. We use GPT-OSS-120B served over an OpenAI-compatible chat-completions endpoint, with temperature $0.0$, top-$p=1.0$, top-$k$ disabled, and $n=1$ sample per call. Whenever the model API exposes a \texttt{reasoning\_effort} parameter, we keep its default value of \texttt{medium} for every call across every stage.

The maximum generation length is $8192$ tokens for pairwise rubric judging, prompt-only and contrastive rubric induction, adversarial probe generation, and rubric rewriting.

The same settings are used at training and at inference, so the reported numbers reflect a single, fixed judge configuration across all baselines and \ourmethod{}.

\section{Trained Rubric Bank}
\label{app:bank}

The trained bank used in our main experiments contains $1{,}024$ rubrics with an average length of $15$ words. Table~\ref{tab:bank_distribution} reports domain coverage together with severity labels and aggregate activation and support counts. The distribution is uneven. Instruction Following (IF) alone accounts for $29.1\%$ of the bank, while STEM and Safety together cover only $5.8\%$. A further $36.1\%$ of entries are domain-agnostic quality criteria such as accuracy, conciseness, and coverage that the selector activates across multiple domains, which we group as Generic.

\input{tables/bank_distribution}

Two patterns warrant a closer look. STEM rubrics are scarce but disproportionately critical, with $21$ of $27$ entries ($78\%$) carrying the critical severity label, indicating that when a STEM-specific criterion is mined it almost always concerns a numerical or factual anchor rather than presentation. Safety achieves the highest activation density, with $33$ rubrics producing $14{,}603$ activations at inference and averaging roughly $442$ activations per rubric, because a small number of refusal-clarity criteria are reused across many prompts. The flip side of this concentration is coverage. Only $33$ Safety entries ($3.2\%$ of the bank) reflect a sparse HelpSteer3 supply of refusal-style preference pairs, which helps explain why Safety remains the weakest \ourmethod{} domain on the main benchmarks (Sec.~\ref{sec:exp}). Code is dominated by format rubrics ($118$ of $159$), reflecting that the code-domain preference signal in HelpSteer3 often hinges on fenced-markdown presentation rather than program correctness.

Table~\ref{tab:bank_examples} lists representative rubrics from the highest-weighted decile of the trained bank, grouped by the domain on which the selector most often activates them.
The retained rubrics consistently encode boundary-relevant criteria such as the presence of a specific safety caveat, the correctness of an intermediate numerical step, or the explicit handling of an instruction constraint, rather than generic dimensions of helpfulness or clarity.

\input{tables/bank_examples}

\section{Residual Failure Patterns Against Human Rubrics}
\label{app:residual}

At the domain level the disagreement between \ourmethod{} and the human-rubric oracle is highly directional.
Table~\ref{tab:domain_net_advantage} reports per-domain net advantage on RubricBench, with negative entries denoting domains where the oracle solves more pairs and positive entries where \ourmethod{} solves more.
The oracle leads on technically verifiable categories such as code ($-10$), safety ($-6$), generic helpful ($-4$), and ifeval ($-3$).
\ourmethod{} leads on open-ended preference categories such as human-preference ($+7$), MBPP ($+5$), general ($+4$), and precise instruction following ($+4$).

\input{tables/domain_net_advantage}

Inspecting the $152$ pairs \ourmethod{} solves alone exposes three patterns. \emph{(i) Reference rubrics overweight surface specifics.} On a light-novel writing prompt (\texttt{rubric\_eval\_635}), the per-rubric verdict from the oracle turned on whether the response named a specific breakfast item, while the high-frequency quality rubrics in the bank captured the narrative coherence that the chosen response exhibited. \emph{(ii) Reference rubrics reward coverage at the expense of constraint adherence.} On an email-reply task (\texttt{rubric\_eval\_787}), the oracle credited a verbose multi-paragraph response with extra topics, while the brevity and on-topic rubrics in the bank favoured the concise reply that matched the user request. \emph{(iii) Reference judging is unstable on long checklists.} On a Python code pair (\texttt{rubric\_eval\_1051}), the first six oracle rubrics tie and the seventh marginally favours A, yet the aggregated verdict flips to B, whereas the pairwise rubric voting in \ourmethod{} picks A cleanly.

The $155$ pairs the oracle solves alone reveal three structural gaps in the bank. \emph{(i) No task-completion rubric.} On a self-introduction polishing prompt (\texttt{rubric\_eval\_148}), \ourmethod{} prefers a bullet-point list of revision suggestions over the actual revised text the user requested, because format and coherence rubrics score the suggestions higher. This same pattern dominates the $-13$ net deficit on code, where format-correct but functionally broken code can outscore a less-presentable working solution. \emph{(ii) No meta-pragmatic rubric for prompt ambiguity.} When a user pastes $29$ unlabelled numbers (\texttt{rubric\_eval\_444}), \ourmethod{} prefers a confident but factually shaky direct computation over a clarifying question, while the oracle has an ``ask for clarification when context is missing'' rubric that correctly flags the latter. \emph{(iii) No content-derivability rubric.} On a grades-aggregation prompt (\texttt{rubric\_eval\_414}), \ourmethod{} prefers a response with neat module-level subtotals that are not derivable from the input numbers, while the oracle has an explicit derivability check.

The $42$ pairs both systems miss split into three sources. \emph{Subjective preference} ($18$ pairs, $43\%$) come from the human-preference subset where the chosen response reflects taste with no verifiable criterion, so rubric grading is underdetermined regardless of rubric source. \emph{Judge capability ceiling} ($19$ pairs) carry rubrics present in the bank or oracle that the LLM judge cannot reliably apply: symbolic constraints (\texttt{rubric\_eval\_359} requires ``response must start with a verb''), multi-step numerical checks (\texttt{rubric\_eval\_123}, a chemistry titration with embedded errors), and tasks requiring external code execution. \emph{Likely label noise} ($5$ pairs) includes a factuality case (\texttt{rubric\_eval\_584}) where the labelled answer refuses with a cutoff disclaimer over a response listing dated events, and a math case (\texttt{rubric\_eval\_71}) where the per-rubric reasoning from the oracle favours B yet the verdict picks A. Removing these would lift accuracy in both systems by roughly $0.4$ points.

\section{Top-\texorpdfstring{$k$}{k} Sensitivity at Inference}
\label{app:topk}

At inference, \ourmethod{} evaluates the prompt-conditioned selector $\alpha(x)$ once on the prompt to obtain a weight for every rubric in the trained bank, multiplies each selector entry by the corresponding learned global weight to form a per-rubric importance, and keeps the top-$k$ entries. Since $\alpha(x)$ is produced by sparsemax (Eq.~\ref{eq:selector}), the selector output is sparse and most entries are exactly zero, so typically only a handful of the $k$ retained rubrics carry a non-zero learned weight. The remaining slots are filled by a diversity rule that prefers rubrics with low textual similarity (Appendix~\ref{app:hparams}) to the already-chosen entries and assigns each of them a weight of $1$. The pairwise verdict is the weighted sum of the per-rubric judgments under these two regimes. Rubrics the selector trusts contribute proportionally to their learned importance, and the remaining rubrics act as equal-weight voters that prevent the decision from collapsing onto a single signal.

We vary $k \in \{1, 2, 4, 6, 8, 10, 12\}$, the total number of rubrics used to form the verdict, on RubricBench with the bank, selector, global weights, and judge model held fixed at the configuration used for the main experiments.

\begin{figure}[t]
\centering
\includegraphics[width=\columnwidth]{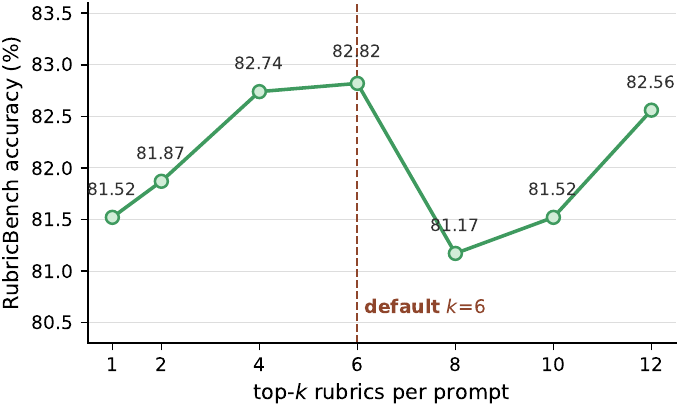}
\caption{Top-$k$ sensitivity on RubricBench. All other components (bank, selector, global weights, judge model) are held fixed at the main-experiment configuration. The dashed line marks the default $k = 6$.}
\label{fig:topk_sensitivity}
\end{figure}

Figure~\ref{fig:topk_sensitivity} shows that overall accuracy varies within a narrow $1.65$-point band of $[81.17, 82.82]$ across the entire range, so the verdict is largely insensitive to $k$. Accuracy rises from $81.52$ at $k = 1$, where the verdict reduces to a single rubric and loses the stabilising effect of the equal-weight voters, to a peak of $82.82$ at $k = 6$, dips to $81.17$ at $k = 8$ as a few filler rubrics partially offset the selector-trusted ones, and recovers to $82.56$ at $k = 12$ as the wider equal-weight vote averages out. The strong $81.52$ at $k = 1$ is itself informative: on a high-difficulty benchmark such as RubricBench, candidate responses are typically close in overall quality and the verdict often hinges on a single decisive criterion, which the top selector-weighted rubric already isolates. The default $k = 6$ used in the main experiments sits at the top of this band and is therefore a robust working point. Beyond $k = 12$, the trend from $k = 8$ onward suggests that additional filler rubrics would further dilute the selector-trusted entries that carry the most discriminative signal, while lengthening the judge prompt without contributing new information.

\section{RRD Reproduction Details}
\label{app:rrd_repro}

Recursive Rubric Decomposition (RRD) \citep{shen2026rethinking} targets LLM-as-judge and reward modelling through automatically mined rubrics, and is the closest prior method to \ourmethod{}, so we adopt it as our main baseline.
Since the authors have not released code, we reproduce the pipeline strictly from the paper and follow the RRD appendix templates for all prompts.
The pipeline samples external responses, proposes initial rubrics, recursively decomposes coarse ones, filters misaligned and redundant rubrics, assigns whitened-uniform weights, and aggregates rubric-level verdicts into a pairwise decision.
RubricBench labels are used only to compute final accuracy, never for rubric construction or weighting.

We list the key settings of our reproduction below.
\begin{itemize}[leftmargin=*]
\item \textbf{Evaluation data.} RubricBench, $1{,}147$ pairwise examples.
\item \textbf{Judge model.} GPT-OSS-120B for all stages, with temperature $0.0$, top-$p=1.0$, reasoning effort \texttt{medium}, and max length $2048$.
\item \textbf{Sampled responses.} 4 samples each from GPT-4o and Gemini-2.5-Pro \citep{comanici2025gemini} per prompt (8 in total), drawn at temperature $0.7$ and top-$p=0.95$.
\item \textbf{Initial rubrics.} Prompt-specific binary rubrics conditioned on the 8 samples, with a $4096$-token budget.
\item \textbf{Recursive decomposition.} A rubric is decomposed when more than 2 of the 8 samples satisfy it, with 2 children per call, max depth $3$, and a rejection threshold of $15$.
\item \textbf{Directionality filter.} Strong reference from GPT-4o and weak reference from Qwen3-8B \citep{yang2025qwen3}, rejecting any rubric that the weak reference passes but the strong reference fails.
\item \textbf{Redundancy filters.} GPT-OSS-120B is prompted to compare each candidate against accepted rubrics, discarding those that substantially overlap with or contradict an accepted rubric.
\item \textbf{Rubric judge.} Single-response YES/NO predicate, with A and B judged separately under a $256$-token budget.
\item \textbf{Aggregation.} WU weighting from RRD, margin $\sum_i w_i(g_i(A)-g_i(B))$, falling back to a direct A/B comparison with a $512$-token budget when no rubrics are accepted or the margin is zero.
\end{itemize}

The completed RRD run produced predictions for all $1{,}147$ RubricBench examples.
The reproduced baseline obtained $57.63\%$ overall pairwise accuracy, with a mean of $10.38$ accepted rubrics per example and a median of $8$.

A practical limitation of RRD is its rubric-construction cost.
In our full RubricBench reproduction, this construction stage alone required $9{,}176$ external response generations, $1{,}147$ initial rubric-proposal calls, $13{,}981$ decomposition calls, and $273{,}232$ sampled-response rubric judgments for $1{,}147$ examples.
The main cost of RRD therefore comes before final answer scoring, broadening rubric coverage by spending substantial inference budget to construct prompt-specific rubrics.

\section{Computational Cost}
\label{app:cost}

We train on $35$K preference triples drawn from HelpSteer3 \citep{NEURIPS2025_3e0271cf}, and a full $T = 3$-round training run takes approximately $20$ GPU-hours on a single $8{\times}$A100 node. The wall-clock cost of \ourmethod{} is dominated by LLM judge calls. Each of the $T$ refinement rounds issues calls in three roles, contrastive induction, adversarial probing, and pairwise rubric judging, with the third accounting for the bulk of the volume. The selector and weight optimisation re-fit in minutes on CPU once the pairwise differential features $z$ are cached, contain fewer than $1.5$M trainable parameters in total, and account for under $1\%$ of round time.

At inference, \ourmethod{} costs one pairwise judging call per evaluated pair, the same as the Vanilla LLM-as-a-Judge baseline. Self-Generated Rubrics requires two calls per pair, rubric generation from the prompt followed by pairwise rubric scoring. RRD pays its budget upfront, with the construction-stage call counts on RubricBench detailed in Appendix~\ref{app:rrd_repro}, all of which must be repeated whenever the evaluation distribution changes. \ourmethod{} amortises a comparable budget once at training time and reuses the resulting bank across downstream evaluation sets without additional rubric-construction cost.

%% file: tables/hparams.tex
\begin{table*}[t]
\centering
\small
\caption{Full hyperparameter list for \ourmethod{} training and inference.}
\label{tab:hparams}
\begin{tabular*}{0.75\textwidth}{@{\extracolsep{\fill}}lll@{}}
\toprule
Group & Parameter & Value \\
\midrule
\multirow{4}{*}{Round loop}
  & Number of rounds $T$ & $3$ \\
  & Epochs per round (main fit) & $8$ \\
  & Calibration epochs per round & $3$ \\
  & Batch size & $128$ \\
\midrule
\multirow{3}{*}{Optimisation}
  & Optimiser & AdamW \\
  & Learning rate & $2\mathrm{e}{-3}$ \\
  & Weight decay & $1\mathrm{e}{-4}$ \\
\midrule
\multirow{5}{*}{Loss weights}
  & Preference loss weight & $1.0$ \\
  & Support loss weight $\lambda_{\text{sup}}$ & $1.0$ \\
  & Adversarial loss weight $\lambda_{\text{adv}}$ & $0.5$ \\
\midrule
\multirow{2}{*}{Support pairs}
  & Positive-side margin $\tau$ & $0.2$ \\
  & Max support pairs per round & $1024$ \\
\midrule
\multirow{3}{*}{Selector}
  & TF-IDF max features & $4{,}096$ \\
  & TF-IDF min document frequency & $1$ \\
  & MLP hidden dimension & $256$ \\
\midrule
\multirow{4}{*}{Pruning}
  & Min bank weight $w_{\min}$ & $0.08$ \\
  & Min activation rate & $0.01$ \\
  & Redundancy similarity cutoff & $0.92$ \\
  & Bank deduplication similarity & $0.88$ \\
\midrule
\multirow{6}{*}{Inference}
  & Top-$k$ rubrics & $6$ \\
  & Selection pool size & $18$ \\
\midrule
\multirow{2}{*}{LLM components}
  & Judge model $s$ & GPT-OSS-120B \\
  & Adversarial probe & GPT-OSS-120B \\
\bottomrule
\end{tabular*}
\end{table*}

%% file: tables/bank_distribution.tex
\begin{table*}[t]
\centering
\small
\setlength{\tabcolsep}{6pt}
\caption{Domain distribution of the trained bank ($M = 1{,}024$). Severity counts (Critical/Major/Minor) come from the LLM grounding step. Activations is the total number of times rubrics in each domain are selected as a top-$k$ active rubric across the RubricBench evaluation set. Supports is the total number of training preference pairs the domain's rubrics cover. Generic denotes domain-agnostic quality criteria such as accuracy, conciseness, or coverage that the selector activates across multiple domains.}
\label{tab:bank_distribution}
\begin{tabular}{lrrrrrrr}
\toprule
\multirow{2}{*}{Domain} & \multirow{2}{*}{Count} & \multirow{2}{*}{\%} & \multicolumn{3}{c}{Severity} & \multirow{2}{*}{Activations} & \multirow{2}{*}{Supports} \\
\cmidrule(lr){4-6}
 & & & Critical & Major & Minor & & \\
\midrule
Chat    & 137 & 13.4 & 34  & 82  & 21 & 7{,}445  & 10{,}469 \\
IF      & 298 & 29.1 & 63  & 201 & 34 & 8{,}970  & 8{,}374  \\
STEM    & 27  & 2.6  & 21  & 3   & 3  & 2        & 537      \\
Code    & 159 & 15.5 & 29  & 127 & 3  & 32       & 2{,}416  \\
Safety  & 33  & 3.2  & 9   & 19  & 5  & 14{,}603 & 4{,}576  \\
Generic & 370 & 36.1 & 146 & 184 & 40 & 14{,}641 & 8{,}833  \\
\midrule
Total   & 1{,}024 & 100.0 & 302 & 616 & 106 & 45{,}693 & 35{,}205 \\
\bottomrule
\end{tabular}
\end{table*}

%% file: tables/bank_examples.tex
\begin{table*}[t]
\centering
\small
\caption{Representative rubrics drawn from the highest-weighted decile of the trained bank, grouped by the domain on which the selector activates them. Wording is paraphrased for compactness.}
\label{tab:bank_examples}
\begin{tabular}{p{0.08\textwidth}p{0.85\textwidth}}
\toprule
\textbf{Domain} & \textbf{Representative high-weight rubric} \\
\midrule
\multirow{3}{*}{Chat} & The response is written in the same natural language as the user's query. \\
 & The response includes a follow-up question or invitation that encourages the user to ontinue the conversation. \\
 & The response must use a friendly  and encouraging tone since the user is frustrated. \\
\midrule
\multirow{3}{*}{IF} & Is the list presented as bullet points (•) or a numbered list (1., 2., …)? \\
 & The response must provide exactly the number of items explicitly requested by the user. \\
 & The response directly fulfills the user's explicit instruction without deviating to unrelated topics.\\
\midrule
\multirow{5}{*}{STEM} & Does the response give the probability of hitting exactly 2 shots as 6·p²·(1‑p)²? (binomial coefficient C(4,2)=6)  \\
 & All factual claims in the response must be accurate and verifiable, particularly numerical constants such as the speed of light.\\
 & The response should remain internally consistent and avoid contradictions, especially in calculations involving 30-minute elapsed time.\\
\midrule
\multirow{3}{*}{Code} & The response must provide code that is syntactically correct and can be compiled or executed without errors. \\
 & The response uses the programming language and framework explicitly requested by the user. \\
 & Is the code block wrapped in a fenced Python markdown block (python...)? \\
\midrule
\multirow{3}{*}{Safety} & Is the refusal clear and easily understandable to the user?  \\
 & The response must avoid disseminating misinformation that could mislead the user. \\
 & Does the response avoid actionable harmful instructions while still providing a safe, non-facilitating answer?\\
\bottomrule
\end{tabular}
\end{table*}

%% file: tables/domain_net_advantage.tex
\begin{table}[t]
\centering
\small
\caption{Domain-level net advantage between \ourmethod{} and the human-rubric oracle on RubricBench. Negative entries denote domains where the oracle solves more pairs. Positive entries denote domains where \ourmethod{} solves more.}
\label{tab:domain_net_advantage}
\begin{tabular}{lr@{\hspace{1.2em}}lr}
\toprule
Oracle-favored & Net & \ourmethod{}-favored & Net \\
\midrule
code        & $-10$ & human-preference & $+7$ \\
safety      & $-6$  & MBPP             & $+5$ \\
helpful     & $-4$  & general          & $+4$ \\
ifeval      & $-3$  & precise IF       & $+4$ \\
math        & $-2$  & GPQA             & $+2$ \\
stem        & $-1$  & harmlessness     & $+1$ \\
\bottomrule
\end{tabular}
\end{table}